\documentclass{article}

\usepackage[preprint]{nips_2018}

\usepackage[utf8]{inputenc} 
\usepackage[T1]{fontenc}    
\usepackage{hyperref}       
\usepackage{url}            
\usepackage{booktabs}       
\usepackage{amsfonts}       
\usepackage{nicefrac}       

 \usepackage[pdftex]{graphicx}
 \usepackage[font=small]{subcaption} 
 \usepackage[font=small]{caption}
 \graphicspath{{../pdf/}{../jpeg/}}
 \DeclareGraphicsExtensions{.pdf,.jpeg,.png}

 \usepackage{amssymb}
 \usepackage{array}
 \usepackage{amsmath}
 \usepackage{graphicx}
 \usepackage{listings}
 \usepackage{amsmath,amssymb}
 \usepackage{amsthm}
 \usepackage{enumerate}
 \usepackage{graphicx}
 \usepackage{times}
 \usepackage{epsfig}
 \usepackage{graphicx}
 \usepackage[font=small]{subcaption} 
 \usepackage[font=small]{caption}
 \usepackage{amsmath}
 \usepackage{amssymb}

 \chardef\other=12

 {\catcode`\|=0 \catcode`|\=\other 
 	|obeylines   
 	|gdef|intt#1^^M{|noalign{#1}}%
 	|gdef|ttfinish#1^^M#2\endtt{#1|let^^M=|cr %
 		|halign{|hskip|parindent|texttt{##}|hfil|cr#2}$$}}
 
 \setcitestyle{square}
 \setcitestyle{numbers}

\title{Visual Attention driven by Convolutional Features}

\author{
  Dario ~Zanca\\
  DINFO, University of Florence\\
  DIISM, University of Siena\\
  \texttt{dario.zanca@unifi.it}
  \And
  Marco ~Gori\\
  DIISM, University of Siena\\
  \texttt{marco@diism.unisi.it}
}

\begin{document}

\maketitle

\begin{abstract}
The understanding of where humans look in a scene is a problem of great interest in visual perception and computer vision. When eye-tracking devices are not a viable option, models of human attention can be used to predict fixations. In this paper we give two contribution. First, we show a model of visual attention that is simply based on deep convolutional neural networks trained for object classification tasks. A method for visualizing saliency maps is defined which is evaluated in a saliency prediction task. Second, we integrate the information of these maps with a bottom-up differential model of eye-movements to simulate visual attention scanpaths. Results on saliency prediction and scores of similarity with human scanpaths demonstrate the effectiveness of this model.
\end{abstract}

\section{Introduction}
High-level cognitive processes such as scene interpretation drammatically benefit of a mechanism of visual attention. This mechanism makes them \textit{tractable}, by focusing resources to one location per time. In humans, visual attention is performed by eye-movements that relocate the fovea to a new spatial location almost four times per second. The problem of formally describing the computational mechanisms underlying this process is of great insterest, both in computational neuroscience and computer vision.
 
In the past three decades, many attempts have been made in the direction of modeling visual attention. Basing on the feature integration theory of attention~\cite{treisman_gelade}, in~\cite{kochull} authors provide a description of human attention that operates in an early representation, which is basically a set of feature maps. They assume that these maps are then combined in a central representation, namely the \textit{saliency map}, which plays a central role in driving the attention mechanisms. A first implementation of this scheme was proposed in~\cite{itti1998}. Several other models have been proposed by the computer vision community, in particular to address the problem of refining saliency maps estimation. For instance, it has been claimed that the attention is driven according to a principle of information maximization~\cite{aim} or by an opportune selection of surprising regions~\cite{bayesiansurprise}. A detailed description of the state of the art is given in~\cite{stateoftheart}. Machine learning approaches have also been used to learn models of saliency. Judd \textit{et al.\@} \cite{Judd} collected 1003 images observed by 15 subjects and trained an SVM classifier with \mbox{low-,} \mbox{middle-,} and \mbox{high-level} features. More recently, automatic feature extraction methods with convolutional neural networks combined with transfer learning achieved top level performance on saliency estimation \cite{sam, deepfix, eDN}.  However, these models do not produce a temporal sequence of eye movements, which can be of great importance for understanding human vision as well as for building systems that deal with video streams. 

Interestingly, recent studies have shown that deep CNNs trained for diverse independent task, exhibit an \textit{inherent} mechanism of spatial selection. Simple convolutional units of different layers of a CNN can actually behave as object detectors, even if no specific supervision is provided for the task~\cite{zhou}. This property has been exploited in different works to perform weakly supervised object localization or related tasks~\cite{cam, gradcam}. These results prove that such models are able to develop the ability of \textit{localising} salient objects without explicit supervision. As a first contribution, in this paper we show that a linear combination of activation maps at the top level of a fully-colvolutional network network trained for the recognition of common objects has competitive results in the task of estimating human fixations distributions.

As a second contribution, we define a model of visual attentive scanpath that relies on the mentioned feature activation maps. Some preliminary attempts to model sequences of fixations (i.e. scanapths) are already present in the literature. Works are often only descriptive \cite{leestella} or task specific \cite{renniger}. In~\cite{lemeur_liu}, authors combine saliency maps with handcrafted biological biases. Others~\cite{cues, hmm} try to combine different feature cues, or even to learn them directly from data~\cite{ming}; but all then evaluate the result with overall statistics (saliency maps refining or average saccade length). A description of visual attention as a dynamic process has been proposed in~\cite {zanca2017}. The authors derive differential equations describing eye-movements, based on three low-level functional principles: boundedness of the retina, curiosity for details and brightness invariance. Dynamic laws are derived within the framework of variational calculus. Functional principle as expressed as energy terms and eye-movements scanpath is naturally obtained as the true trajectory of a mass $m$ in the defined mechanical system. This approach exhibits the important advantage of avoiding the global computation of a saliency map in advance, making it suitable for real-time applications. In this paper we extend this model with the top-down semantic information captured by deep convolutional neural networks pre-trained for the classification of common objects. This information is added in a natural way as fourth functional component, thus demonstrating the versatility of the framework for possible task-oriented experiments. To stress the model, we used a wide collection of images including basic feature (pattern, sketch, fractals), noisy and low resolution, natural landscape, abstract (cartoon and line drawing), high level semantic content (social, affective, indoor), and more. The model is evaluated on the task of saliency prediction. Furthermore, similarities between simulated scanpaths and human scanpaths are measured with different metrics.

The paper is organized as follow. In section \ref{semanticmaps}, we describe how to obtain activation maps from the last convolutional layer of a fully-convolutional CCN. Activation maps from inception-v3~\cite{inceptionv3} are evaluated as saliency predictors. In section \ref{integrating}, the model described in~\cite{zanca2017} is extended with the information carried by the activation maps. Results for both saliency prediction and for scanpath similarity prove the effectiveness of the model. Section \ref{conclusion} contains conclusions and suggestions for future works.

\section{Inherent visual attention in deep convolutional neural networks}\label{semanticmaps}
Recently, in the strand of explainability of deep learning, efforts have been made to understand what deep models \textit{actually} learn. In the case of CNNs, some methods~\cite{cam, gradcam} allow  to visualize internal activation and understand which locations of the original input were crucial for the system response. In particular, authors remove the fully-connected layer before the final output and replace it with global average pooling followed
by a fully-connected softmax layer. Class-specific activation maps are then obtained by averaging feature maps from the last convolutional tensor with the weights of the correspondent class. 

Guided by the same principle but not being interested in one particular class, we claim that  semantic maps obtained by averaging the activation of the units in the last convolutional layer are good predictors for human fixations distribution. In this section, we make a  quantitative analysis of how well this maps predict human fixations, even if they have been trained for a different task.

\begin{figure}[t]
	\begin{center}
		\begin{subfigure}[t]{.49\linewidth}
			\includegraphics[width=.98\linewidth]{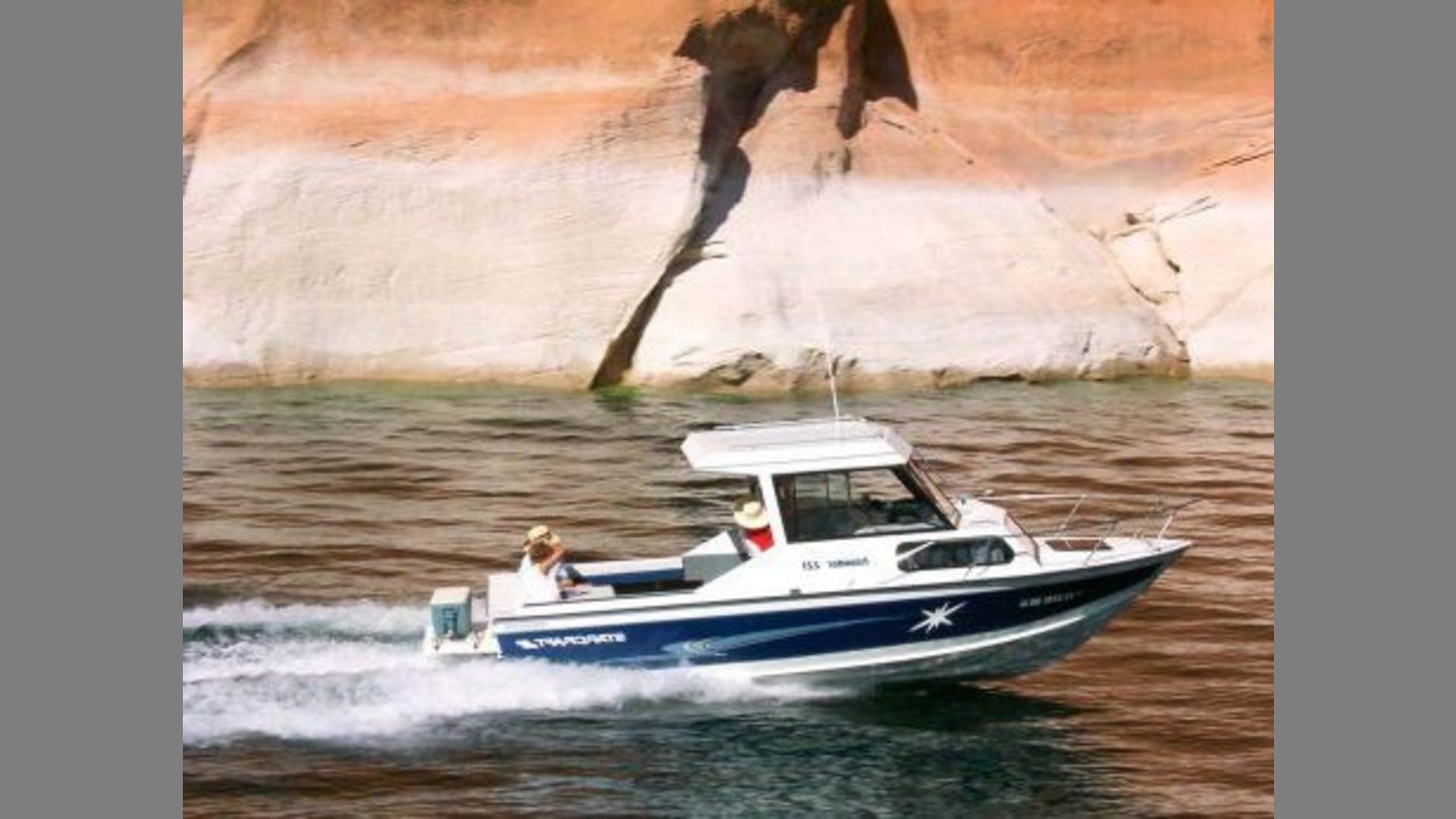}
		\end{subfigure}
		\begin{subfigure}[t]{.49\linewidth}
			\includegraphics[width=.98\linewidth]{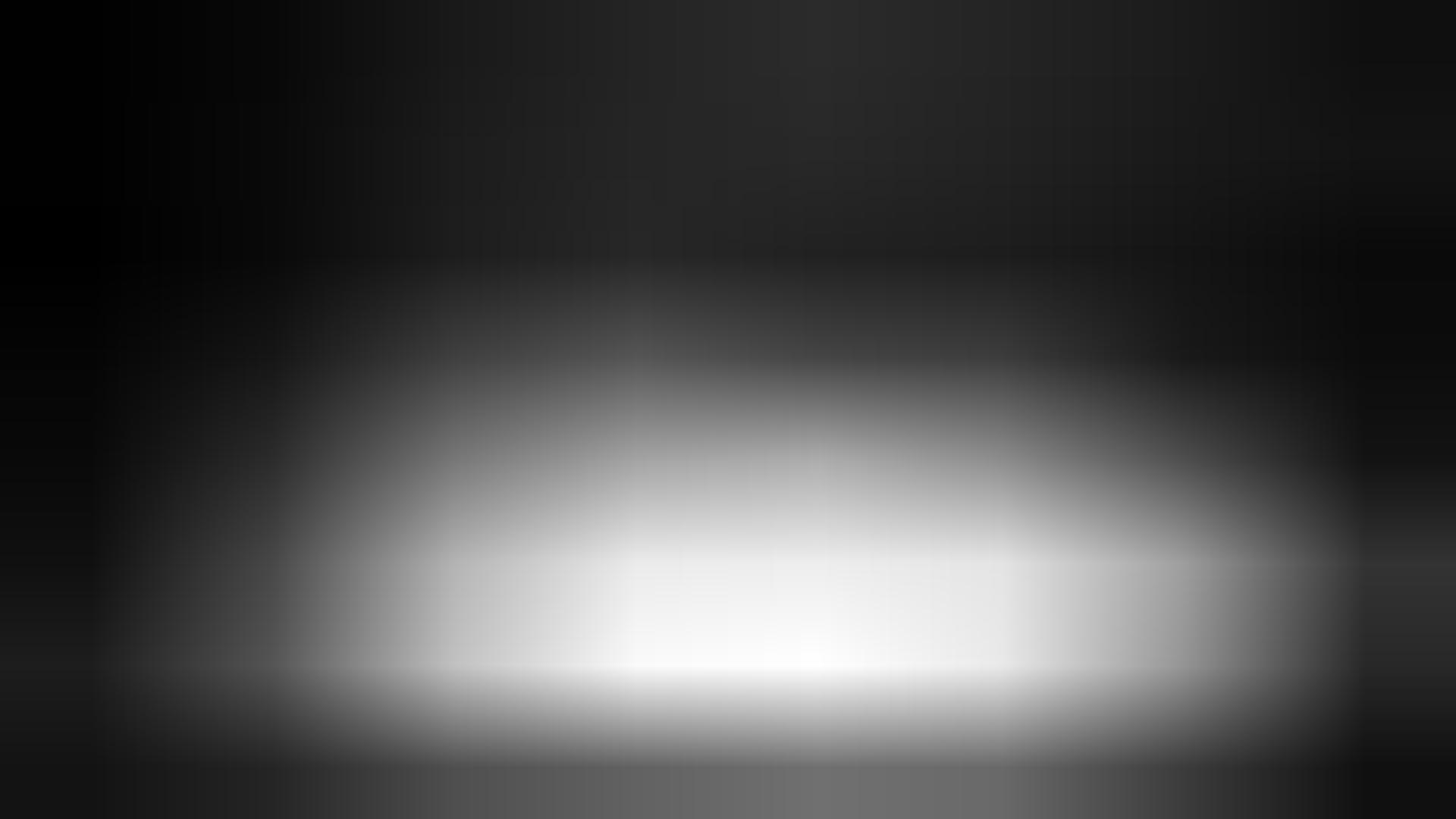}
        \vspace{.5em}
		\end{subfigure}	
		\begin{subfigure}[t]{.49\linewidth}
			\includegraphics[width=.98\linewidth]{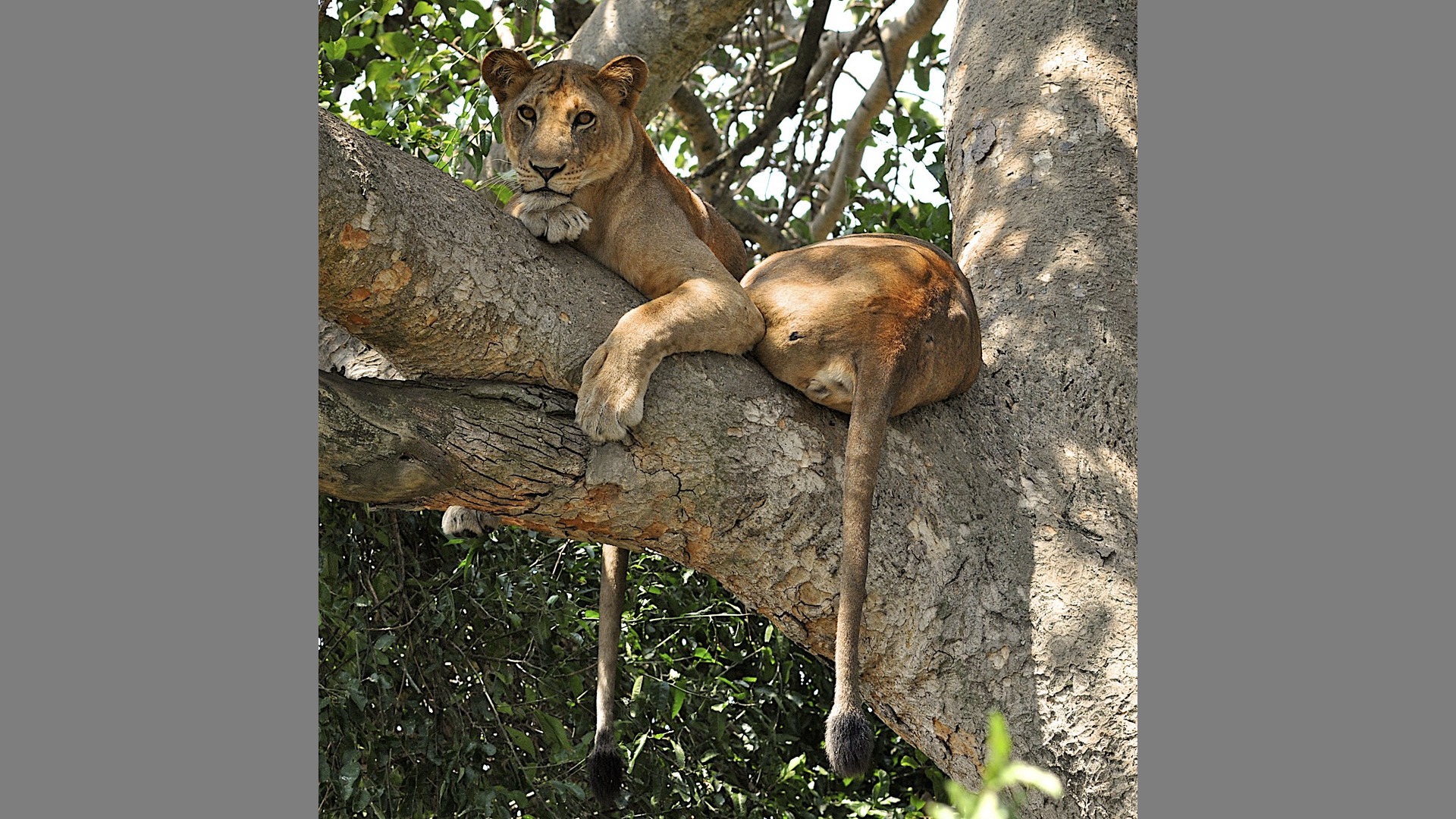}
			\caption{Stimulus}
			\label{fig:stimulus}
		\end{subfigure}
		\begin{subfigure}[t]{.49\linewidth}
			\includegraphics[width=.98\linewidth]{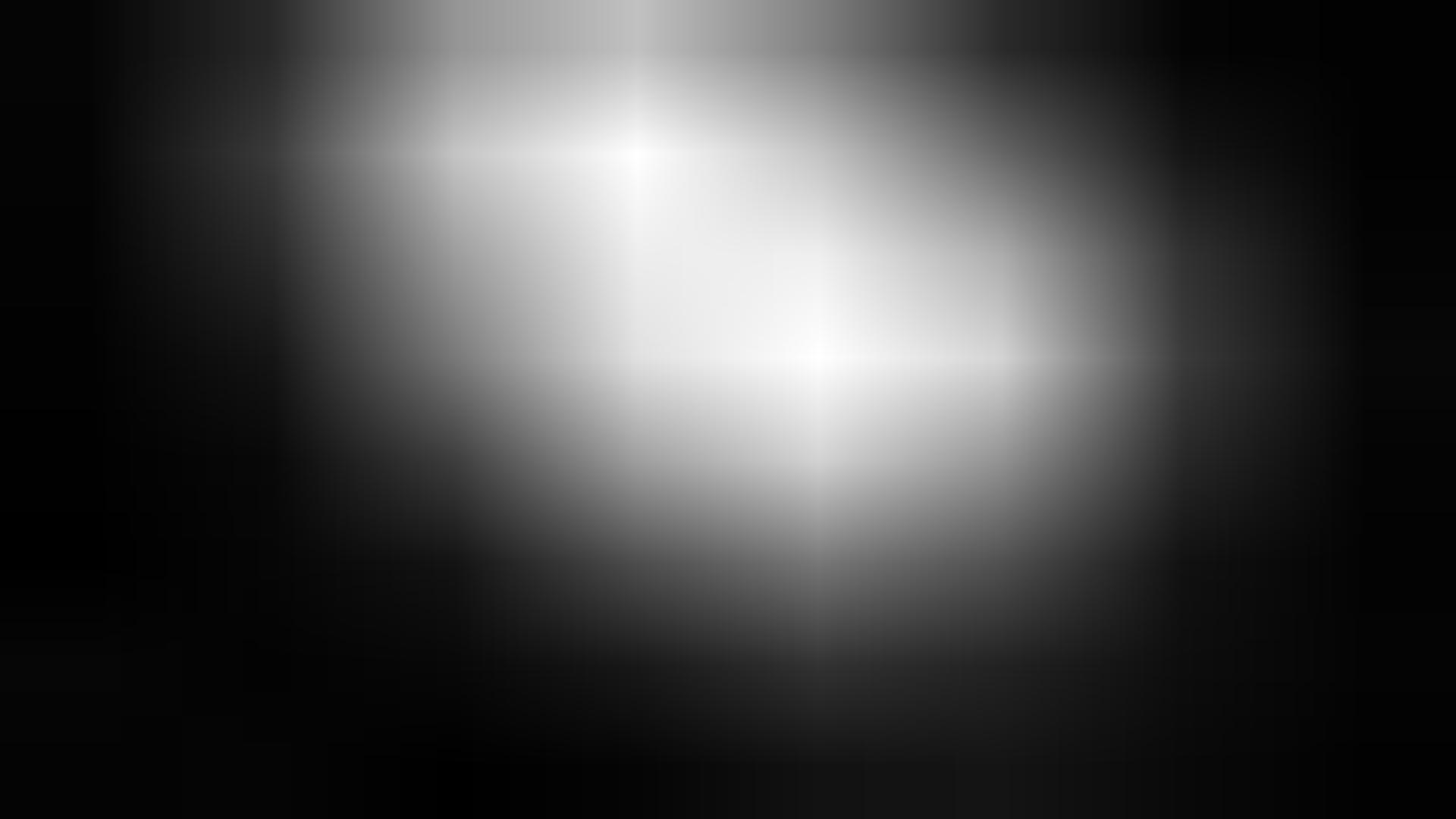}
			\caption{$M$}
			\label{fig:lcta}
		\end{subfigure}	
	\end{center}
	\caption{\textbf{Convolutional feature activation map M.} In column \ref{fig:stimulus}, examples of images from CAT2000~\cite{cat2000}. In column \ref{fig:lcta} the correspondent map $M$ obtained from the pre-trained instance of inception-v3~\cite{inceptionv3}.}
	\label{fig:lctaexample}
\end{figure}

\begin{figure}
    \centering
    \includegraphics[width=.49\linewidth]{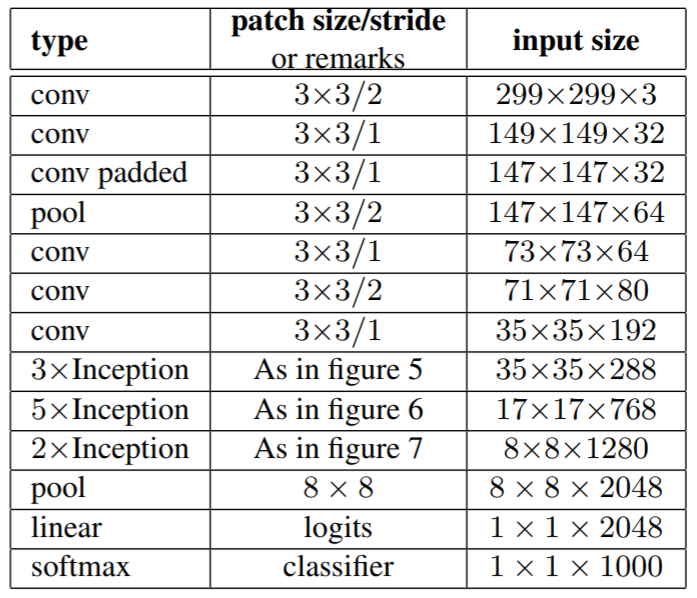}
    \caption{\textbf{Inception-v3.} Architecture specifications of the model of CNN described in~\cite{inceptionv3}.}
    \label{fig:inception-v3-specs}
\end{figure}

In our experiments, we used an instance of the model described in~\cite{inceptionv3}, pre-trained for classification on the ImageNet benchmark~\footnote{http://image-net.org}. The architecture is described in detail in Fig.~\ref{fig:inception-v3-specs}. Since it is a fully-convolutional model (until the last level before the final softmax) it is suitable for the purpose of constructing semantic maps~\cite{cam}.

For a given input image, let $f_k(x)$ represents the activation of unit $k$ in the last convolutional layer "pool" at spatial location $x = (x_1, x_2)$. Then, we can indicate the result of global average pooling as 
\begin{equation}
F_k = \sum_{x} f_k(x).
\end{equation}

For each class $c$ of the dataset, the input of the softmax is 
\begin{equation}
\sum_k w_k^c F_k,
\end{equation} 
where $w_k^c$ is the weight corresponding to class $c$ for unit $k$. Notice that $w_k^c$ expresses the importance of $F_k$ for the class $c$. In~\cite{cam}, class-specific activation maps are defined for each class as
\begin{equation}
M_c(x) = \sum_k w_k^c f_k(x).
\end{equation}

For our purpose of building a model  free-viewing, we are not interested in any specific class. Then, we can average the activations by setting 
\begin{equation}
w_k^c = k^{-1}, \forall k.
\end{equation}

We remove the subscript $c$ and indicate with 
\begin{equation} \label{M}
    M(x) = \frac{1}{k} \sum_k f_k(x)
\end{equation}
the map of the average activations of the features on the last convolutional layer, defined on each spatial location $x$. Examples of this maps are given in Fig.~\ref{fig:lctaexample}. Notice that the resulting maps have a much lower resolution than the original input, because of the many pooling operations. In our experiments, these maps have been resized with cubic interpolation to the original input size in order to obtain a map $M$ defined on each pixel. 

The regions of the image with the highest activation are those more likely to contain the most relevant information for the task, i.e. the most salient object of the scene. If the hypothesis that humans direct their gaze towards objects is true~\cite{peronaobject}, we may expect to find correlation between these maps and the humans fixations maps. We have quantitatively evaluated this hypothesis on CAT2000~\cite{cat2000} by measuring how well $M$ predicts human fixations maps. Performance improve by optimizing maps with blurring and histogram matching~\cite{whatmetrics}. Scores are reported in Tab.~\ref{tab:scoreCF}. From now on, we will refer to this model as the Convolutional Feature map (CF).

\begin{table*}[h]
\begin{center}
	\begin{tabular}{|l l|c c|}
		\hline
		\multicolumn{2}{|c|}{} & \multicolumn{2}{|c|}{CAT2000}\\
		Model version & Maps optimization & AUC & NSS \\
		\hline\hline			
		{\small CF} & - & 0.80 (0.001) & 1.177 (0.046) \\
		\hline
		{\small CF} & center bias	 & \textbf{0.844 (0.001)} & 1.168 (0.009) \\
		\hline				
		{\small CF} & center bias, histogram matching	 & 0.834 (0.001) & \textbf{1.684 (0.085)} \\
		\hline\hline			
		{\small Itti-Koch}&{\cite{itti1998}, implem. by~\cite{harel}} &  0.77 & 1.06\\
		\hline			
		{\small AIM}&{\cite{aim}}  & 0.76 & 0.89  \\
		\hline		
		{\small Judd Model}&{\cite{Judd} }  & 0.84 & 1.30 \\
		\hline			
		{\small AWS}&{\cite{aws}}  & 0.76 & 1.09\\
		\hline
		{\small eDN}&{\cite{eDN}}  & 0.85 & 1.30 \\
		\hline		
		{\small DeepFix}&{\cite{deepfix}}  & 0.87 & 2.28 \\
		\hline		
		{\small SAM}&{\cite{sam}}  & \textbf{0.88} & \textbf{2.38} \\
		\hline						
	\end{tabular}
\end{center}
\vspace{-0.5em}
\caption{Results on CAT2000~\cite{cat2000}. Between brackets is indicated the standard error.}\label{tab:scoreCF}
\end{table*}

\section{The scanpath model}\label{integrating}

\paragraph{Bottom-up mechanism}
In~\cite{zanca2017}, a dynamic model of visual attention is derived by three functional principles. First, eye movements are required to be \textit{bounded inside the definite area of the retina}, 
    \begin{equation} \label{retina}
    	V(x) = k \sum_{i=1,2} \big( \left( l_i - x_i\right)^2 \cdot \left[ x_i > l_i \right] + \left( x_i \right)^2 \cdot \left[ x_i < 0 \right] \big).
    \end{equation}
Second, locations with high values of the \textit{brightness gradient are attractive}. This gives the potential term
    \begin{equation}\label{curiosity}
     	C(t, x) =  b_x^2 \cos^2(\omega t) 
    	+ p_x^2 \sin^2(\omega t).
    \end{equation}
Finally, trajectories are required to preserve the property of \textit{brightness invariance}, which brings to fixation and tracking behaviors. This is guaranteed by the soft satisfaction of the constraint
    \begin{equation}\label{bi}
    	B(t,x, \dot x) = \big(b_t + b_{x} \dot{x}\big)^{2}
    \end{equation}
to be as small as possible. This makes it possible to construct the generalized action
    \begin{equation} \label{S}
    S = \int^T_0 L(t, x, \dot x) \,dt
    \end{equation}
    where $L=K-U$,  where $K$ is the kinetic energy
    \begin{equation} \label{K}
    K(\dot x)= \frac{1}{2}m \dot{x}^2
    \end{equation}
    and $U$ is a generalized potential energy defined as 
    \begin{equation} \label{U}
    U(t, x, \dot x) = V(x) - \eta C(t, x) + \lambda B(t, x, \dot x).
    \end{equation}
Notice that, while $V$ and $B$ get the usual sign of potentials, $C$ comes with negative
sign. This is because, differently from the other two terms, $C$ generates an attractive field. Moreover, it is worth mentioning that term in~\ref{bi} is not a truly potential,
since it depends on both the position and the velocity. However, its generalized interpretation
comes from considering that it generates a force field. Results on stationarity of the solution still hold in this case~\cite{GelfandFomin}.

By the Principle of Least Action, the true path of a mass $m$ within the defined potential fields is given by the Euler-Lagrange equations
	\begin{equation}\label{equations_of_motion_general_form}
	m \ddot{x}- \lambda  \frac{d}{dt} B_{\dot{x}} +
	V_{x} - \eta C_{x} + \lambda 
	B_{x}  = 0.
	\end{equation}

Authors in~\cite{zanca2017} indicates~\ref{equations_of_motion_general_form} as the  Eye Movements Laws (EYMOL). This equations can be numerically integrated to simulate processes of free visual exploration. 

\paragraph{Top-down signal integration}
Nevertheless, the model in~\cite{zanca2017} is too naive. It fails in those categories that contain high level semantic content, for example those picture containing faces, writings, emotional content. The principles~\ref{retina},~\ref{curiosity} and~\ref{bi} are, in fact, very local: they depend on the fixated pixel value and its small surround and fail in capturing properties on the object-level.

This can be solved by adding to the system a top-down signal which suggest which regions are more likely to contain an object and, even more, the main object of the scene. For this reason, we extend the model in~\cite{zanca2017} by adding the information carried by~\ref{M}. The versatility of the framework allows us to do so simply by modifying the potential energy~\ref{U} as follows
    \begin{equation}\label{newU}
        \bar U(t, x, \dot x) = U(t, x, \dot x)- \gamma M(x),
    \end{equation}
that brings to the new model
	\begin{equation}\label{new_equations}
	m \ddot{x}- \lambda  \frac{d}{dt} B_{\dot{x}} +
	V_{x} - \eta C_{x} + \lambda 
	B_{x} - \gamma M_{x}  = 0.
	\end{equation}
	
Where $M$ corresponds to the term define in~\ref{M} and $M_x$ to its spatial derivative. The top-down signal $M$ carries the information of how much the pixel $x=(x_1, x_2)$ belongs to a salient object. From now on, we will refer to the EYMOL model enriched with convolutional features CF with the acronym CF-EYMOL.

Notice that, this method for integrating a top-down signal is very general. The top-down signal is calculated independently of the eye movement. It may provide information other than the preference of object-like figures. For example, the mechanism may be asked to prefer eye movements that minimize the distance to a certain target (tracking task), or it may be asked to favor fixations on those locations that contain a certain feature of interest (search task).

\section{Experimental results}

\paragraph{Saliency prediction} Differently by many of the most popular methodologies in the state-of-the-art, the saliency map is not itself the central component of our model. The output of the model is a trajectory determined by a system of two second ordered differential equations~\ref{new_equations}, provided with a set of initial conditions. Saliency map is calculated as byproduct by summing up the most visited locations after a certain number of virtual observations. Tab.~\ref{tab:score199} reports scores for saliency maps obtained by running the model 199 times, each run was randomly initialized almost at the center of the image and with a small random velocity, and integrated for a running time corresponding to 1 second of visual exploration. The addition of convolutional features CF brings improvements in performance.

\begin{table*}[h]
\begin{center}
	\begin{tabular}{|l l|c c|}
		\hline
		\multicolumn{2}{|c|}{} & \multicolumn{2}{c|}{CAT2000}\\
		Model version & Maps optimization & AUC & NSS \\
		\hline\hline			
		{\small EYMOL} & blur & 0.838 (0.001) & 1.810 (0.014) \\
		\hline
		{\small CF-EYMOL} & blur	 & \textbf{0.843 (0.001)} & \textbf{1.822 (0.064)} \\
		\hline\hline			
		{\small Itti-Koch}&{\cite{itti1998}, implem. by~\cite{harel}} &  0.77 & 1.06\\
		\hline			
		{\small AIM}&{\cite{aim}}  & 0.76 & 0.89  \\
		\hline		
		{\small Judd Model}&{\cite{Judd} }  & 0.84 & 1.30 \\
		\hline			
		{\small AWS}&{\cite{aws}}  & 0.76 & 1.09\\
		\hline
		{\small eDN}&{\cite{eDN}}  & 0.85 & 1.30 \\
		\hline		
		{\small DeepFix}&{\cite{sam}}  & 0.87 & 2.28 \\
		\hline		
		{\small SAM}&{\cite{sam}}  & \textbf{0.88} & \textbf{2.38} \\
		\hline				
	\end{tabular}
\end{center}
\vspace{-0.5em}
\caption{Results on CAT2000~\cite{cat2000} for 199 observations. Between brackets is indicated the standard error.}\label{tab:score199}
\end{table*}

Improvements are more evident in the case in which the time of image exploration is dramatically reduced from $199$ to $10$ seconds, as it can be seen in Tab.~\ref{tab:score10}. This can be explained by the fact that the peripheral prior provided by the convolutional features CF is more crucial when only a few fixations are granted to the system. It is worth notice that, this last configuration runs in real-time in a average equipped personal computer and its performance is comparable or better than the \textit{one-human} baseline.

\begin{table*}[h]
\begin{center}
	\begin{tabular}{|l l|c c|}
		\hline
		\multicolumn{2}{|c|}{} & \multicolumn{2}{c|}{CAT2000}\\
		Model version & Maps optimization & AUC & NSS \\
		\hline\hline			
		{\small EYMOL} & blur & 0.805 (0.001) & 1.428 (0.031) \\
		\hline
		{\small CF-EYMOL} & blur	 & \textbf{0.821 (0.001)} & 1.524 (0.040) \\
		\hline				
		{\small \{Baseline: one-human\}}&\cite{mitsaliencybenchmark}	 & 0.76 & \textbf{1.54} \\
		\hline				
	\end{tabular}
\end{center}
\vspace{-0.5em}
\caption{Results on CAT2000~\cite{cat2000} for 10 observations. Between brackets is indicated the standard error.}\label{tab:score10}
\end{table*}

\paragraph{Scanpath Similarity} Most state-of-the-art computational models in visual attention estimate the probability distribution of fixating a certain image location, i.e. the saliency map~\cite{Judd, aim, cam, deepfix}. However,  these models do not produce a temporal sequence of eye movements, which can be of great importance for understanding human vision as well as for building  systems that deal with video streams. Some efforts have been made in order to model sequences of fixations (i.e. scanapths) but they are often only descriptive \cite{leestella}, task specific \cite{renniger} or evaluate the result with overall statistics like saliency map refinement or average saccade length~\cite{lemeur_liu, cues, hmm, ming}.

In order to evaluate the behavioral properties of our model, we compute the distance (or similarity) between simulated scanpaths and human scanpaths. In the neuroscience literature, the two main metrics proposed to measure the distance (or similarity) between two sequences of fixations are:
\begin{itemize}
	\item \textit{"String-edit" or Levenshtein distance (distance).} The algorithm of the dynamic program to compute this metric is
	described in~\cite{Jurafsky}. The metric has been used in different works for comparing different human
	scanpaths~\cite{Brandt, Foulsham}. It has been shown~\cite{Choi} to be robust to changes in the number of regions
	used to divide the input stimulus. In particular, input stimulus is divided into $n \times n$ regions, labeled with characters. Scanpaths are turned into strings by associating the corresponding character to each fixation. Finally, the algorithm of string-edit is used to measure the distance between the two generated strings.
	\item \textit{Scaled time-delay embedding (similarity).} Time-delay embedding are used in	order to quantitatively compare stochastic and dynamic scanpaths of varied lengths. This metric is largely used in dynamic systems analysis and carefully described in~\cite{Wang}. In our experiment we used a scaled version~\cite{FixaTons}, which is invariant with respect of the size of the input image and average above all possible sub-sequence lengths.
\end{itemize}
Since the output of the proposed model produces a simulated continuous trajectory, it is necessary to extract fixations from it in order to compute the defined metrics. For this purpose, we use the standard python library pygaze~\footnote{http://www.pygaze.org/2015/06/pygaze-analyser/} to extract fixations from raw data of the eye-tracker device. Threshold values have been left unchanged to extract only human-like fixations. Tab.~\ref{tab:simscore} illustrates the experimental results for the collection of the four different datasets MIT1003~\cite{Judd}, SIENA12~\cite{FixaTons}, TORONTO~\cite{aim}, KOOTSTRA~\cite{kootstra}. The collection includes a total of 1233 images of different sizes and semantic categories. The model is compared with two baselines:
\begin{itemize}
	\item \textit{Random.} Fixations are sampled from a uniform distribution.
	\item \textit{Center.} Fixations are sampled from a gaussian distribution, centered in the middle of the image~\cite{Judd, mitsaliencybenchmark}.
\end{itemize}

\begin{table*}[h]
	\begin{center}
		\begin{tabular}{|l|c c | c c|}
			\hline
			 &\multicolumn{4}{ c|}{\textbf{DATASET COLLECTION}~\cite{Judd,FixaTons,aim,kootstra}}\\
			 \hline
			  & \multicolumn{2}{c|}{\textbf{String-Edit (distance)}} &  \multicolumn{2}{c|}{\textbf{Scaled Time-delay embedding (similarity)}} \\
			\textbf{Model}&\textbf{Average}&\textbf{Best}&\textbf{Average}&\textbf{Best}\\
			\hline\hline			
			CF-EYMOL&9.36 (4.08)& \textbf{3.75 (2.78)}& \textbf{0.822 (0.001)} & \textbf{0.917 (0.001)}\\
			\hline
			\{Random\}&9.29 (2.70)&7.67 (0.76)&0.737 (0.001)&0.770 (0.001)\\
			\hline
			\{Center\}&9.28 (2.84)&7.68 (0.87)&0.724 (0.002)&0.762 (0.002)\\	
			\hline				
		\end{tabular}
	\end{center}
	\vspace{-0.5em}
	\caption{Results on a collection of four datasets: MIT1003~\cite{Judd}, SIENA12~\cite{FixaTons}, TORONTO~\cite{aim}, KOOTSTRA~\cite{kootstra}. Models between curly brackets are baseline. Between brackets is indicated the standard error. For each model, 10 differrent scanpaths are simulated. Average is calculated as the mean over all simulated scanpaths. Best is calculated considering, for each input stimulus, only the scanpath with the best score. The best scores are highlighted in bold.}\label{tab:simscore}
\end{table*}

The results show that the baseline \textit{center} is only slightly better than the \textit{random} baseline. Our CF-EYMOL model is significantly better than the two baseline for both metrics. Notice that, especially in the \textit{Best} case, which considers only the best score for each image, simulated scanpaths are very closely correlated to human scanpaths.

\begin{figure}
	\begin{center}
		\begin{subfigure}[t]{.49\linewidth}
			\includegraphics[width=.98\linewidth]{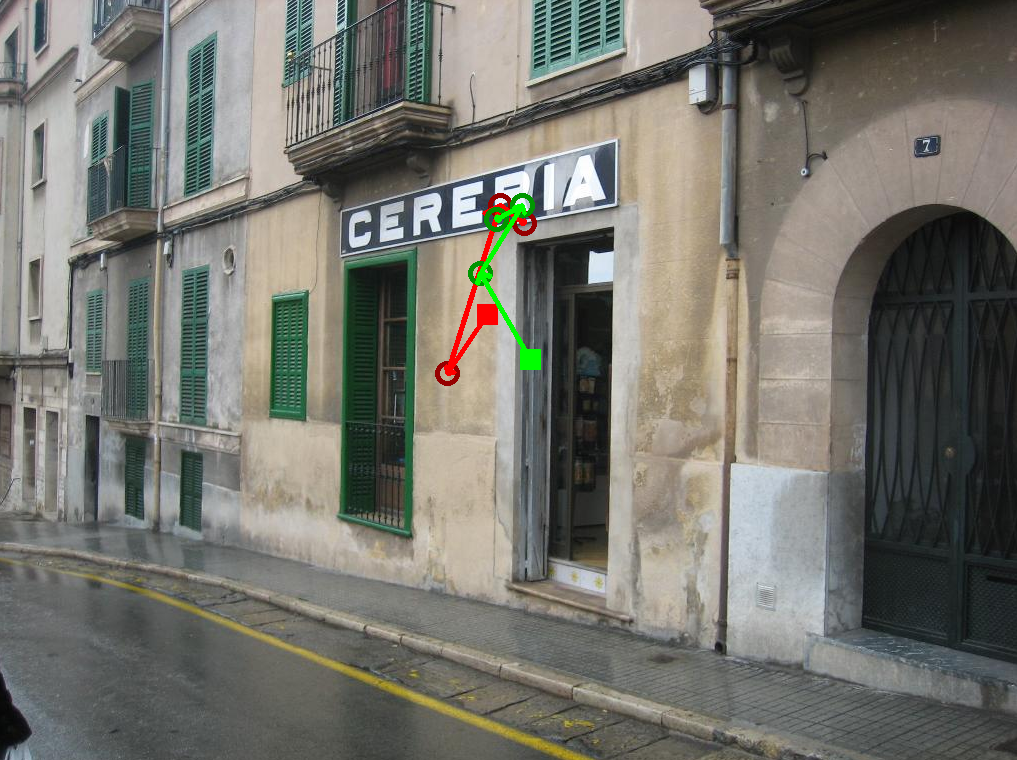}
		\end{subfigure}
		\begin{subfigure}[t]{.49\linewidth}
			\includegraphics[width=.98\linewidth]{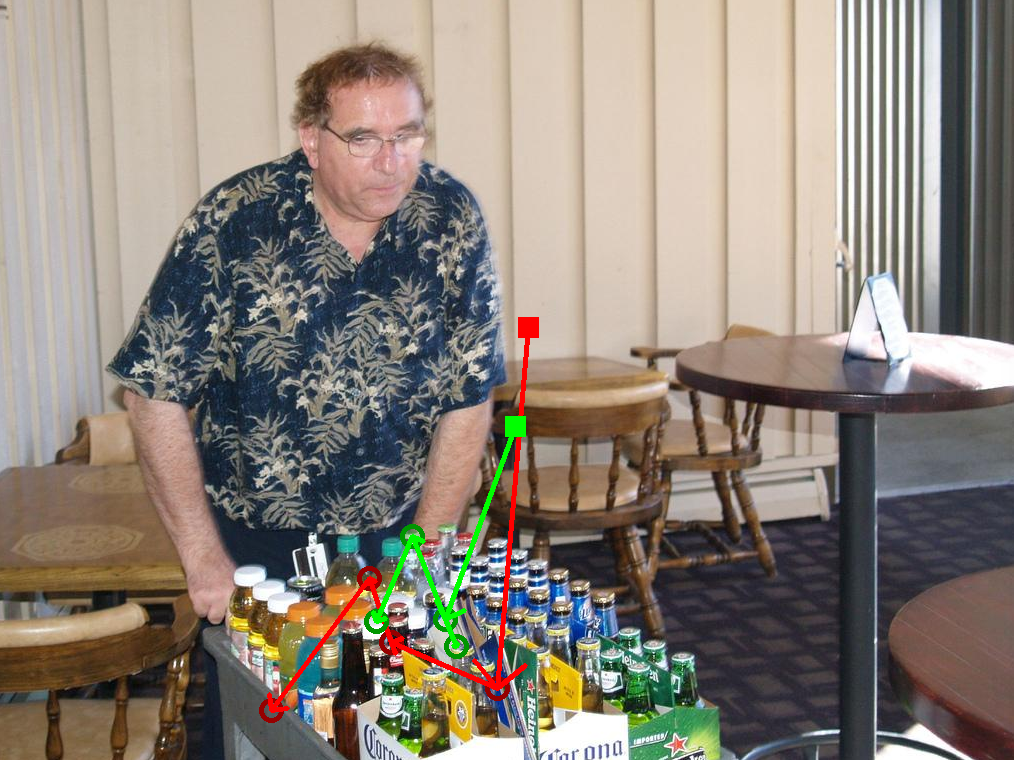}
			\vspace{.5em}
		\end{subfigure}
		\begin{subfigure}[t]{.49\linewidth}
			\includegraphics[width=.98\linewidth]{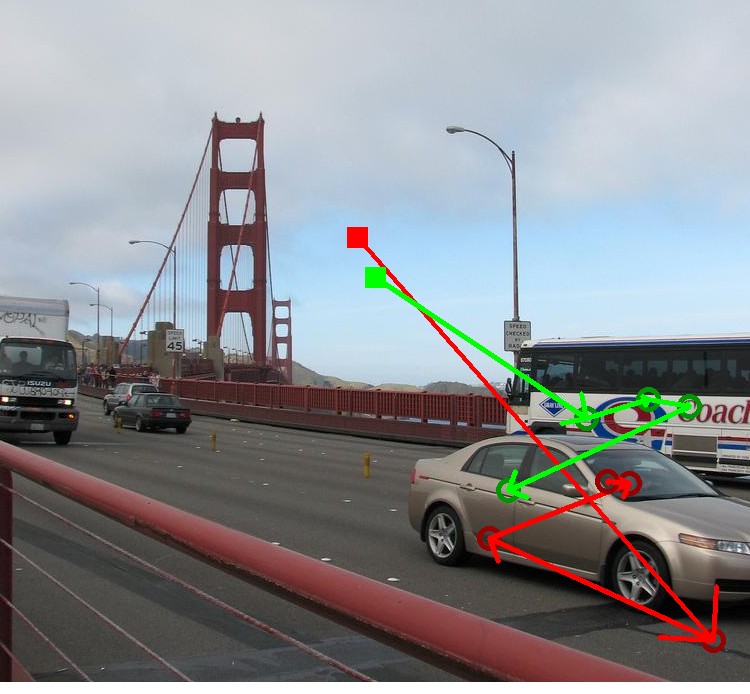}
		\end{subfigure}
		\begin{subfigure}[t]{.49\linewidth}
			\includegraphics[width=.98\linewidth]{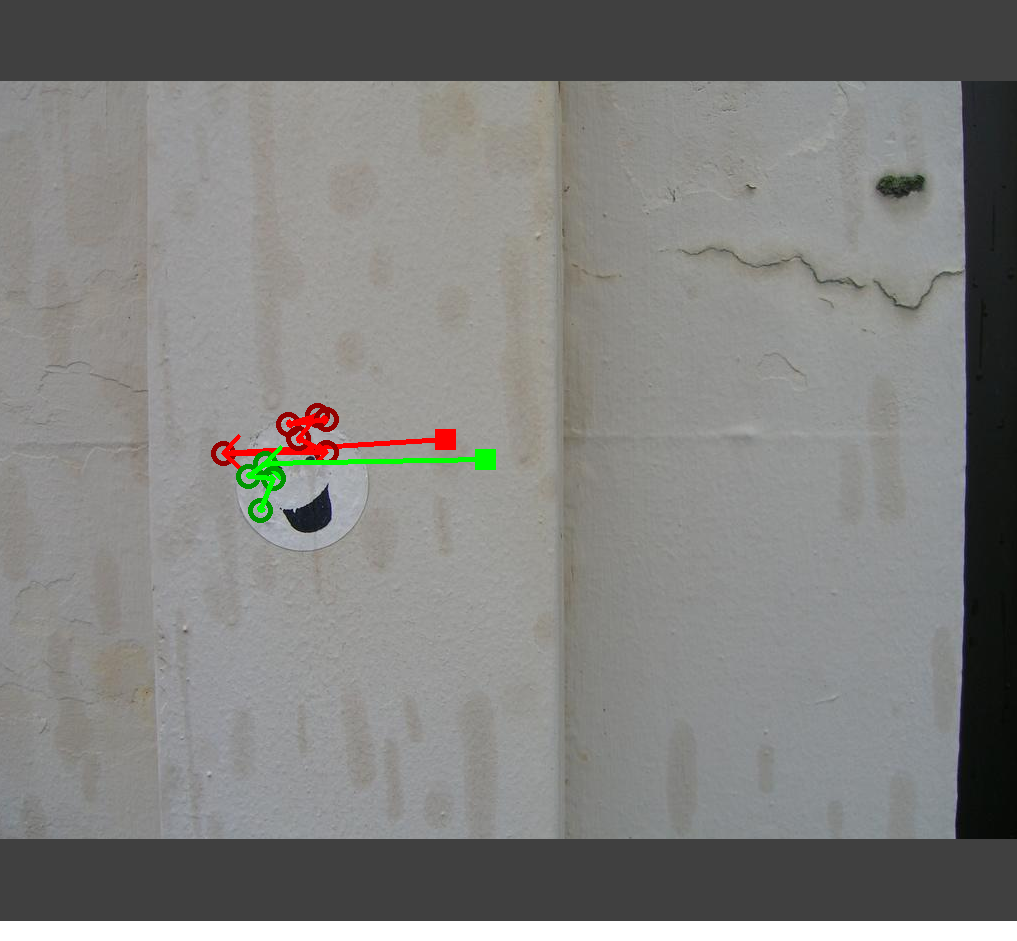}
		\end{subfigure}		
		\begin{subfigure}[t]{.49\linewidth}
			\includegraphics[width=.98\linewidth]{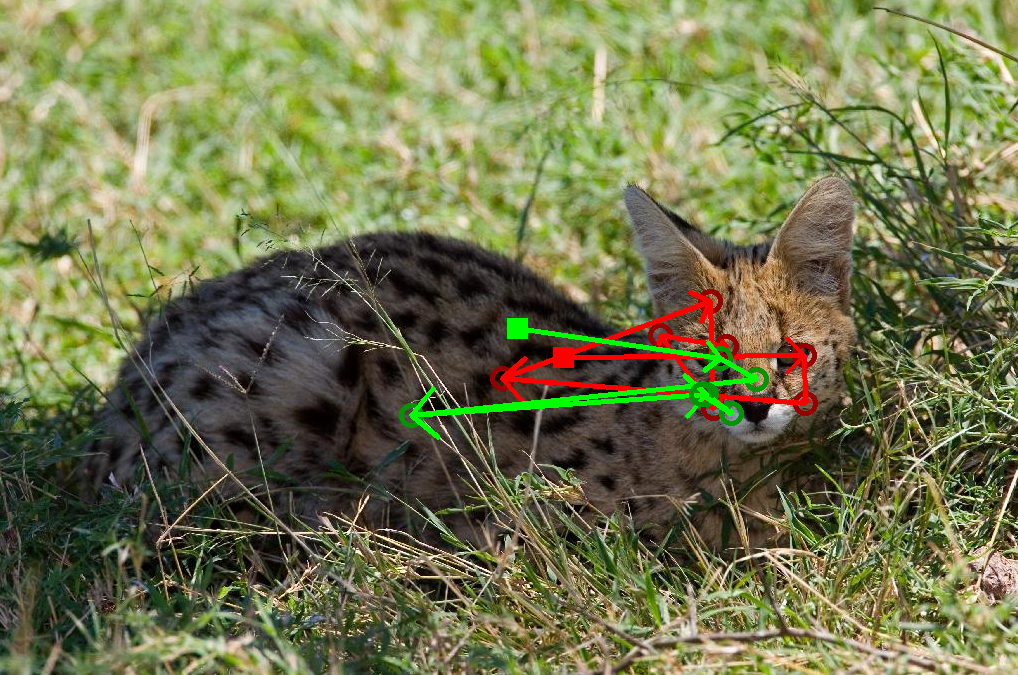}
		\end{subfigure}
		\begin{subfigure}[t]{.49\linewidth}
			\includegraphics[width=.98\linewidth]{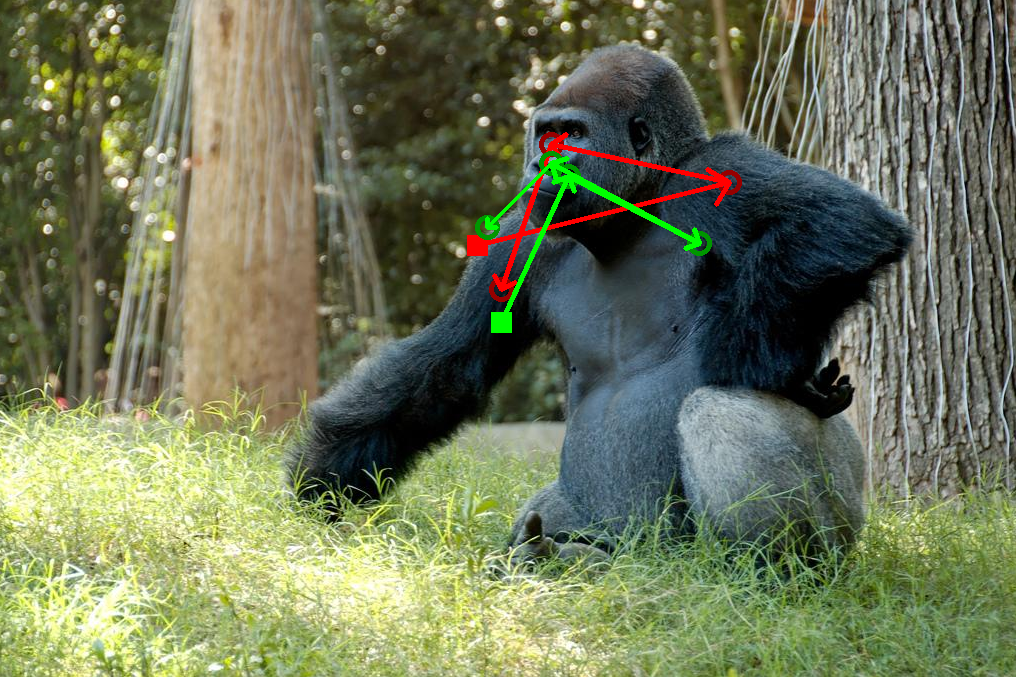}
		\end{subfigure}	
	\end{center}
	\caption{\textbf{Examples of simulated scanpaths.} This figure shows some of the best cases according to the similarity metrics used. The generated sequences of fixations  are compared with the nearest human scanpath. Simulated scanpaths are drawn in red, human scanpath in green. The starting point is indicated by a square and the direction of the saccades by an arrow.} 
	\label{fig:scanpaths}
\end{figure}

Fig.~\ref{fig:scanpaths} shows some examples of simulated scanpaths compared to human scanpaths. The examples reported are chosen among those that produce the best scores according to the defined metrics. Notice that in many cases the trajectories start from the same point and follow the same direction. The simulated trajectories, due to the addition of convolutional CF features, are strongly attracted by objects (car, text, faces).

\section{Conclusion and future works}\label{conclusion}

In this paper we have shown that  an inherent capability of modeling visual attention is present in deep convolutional neural networks that are trained for a different task (i.e., object classification). The experimental results provide evidence that the convolutional features leads good human saliency predictors. However,  the main contribution of this paper is to integrate the information brought by these maps with the bottom-up differential model of eye-movements defined in~\cite{zanca2017}, with the final purpose of  simulating visual attention scanpaths. The proposed integration enriches the eye movement model thanks to the additional peripheral information that comes from the convolutional filters.

The results in saliency prediction show that the model competes with state-of-the-art models and that it is particularly effective in the case in which observation time is drastically limited (10 seconds). This is very promising, especially if we consider that the proposed theory offers a truly model of eye movements, whereas the computation of the saliency maps only arises as a byproduct. 

A correct evaluation of a scanpath model must include some behavioral measures of scanpath similarity. For this purpose, in this paper we have presented two suitable metrics for dynamic processes and calculated two baselines. The scanpath similarity measures demonstrate that the  scanpaths simulated with the proposed model are very correlated with human scanpaths. This result deserves further scientific investigation to actually verify how much the differential model captures oculomotor aspects of humans eyes.

\end{document}